\documentclass[conference]{IEEEtran}
\usepackage{cite}
\usepackage{amsmath,amssymb,amsfonts}
\usepackage{algorithmic}
\usepackage{graphicx}
\usepackage{textcomp}
\usepackage{xcolor}
\usepackage{booktabs}
\usepackage{float}
\usepackage{hyperref}
\usepackage{academicons}
\usepackage{hyperref}
\usepackage{xcolor}
\usepackage{fontawesome5}

\definecolor{orcidgreen}{HTML}{A6CE39}
\begin{document}

\title{Empowering Local Agriculture: A Deep Learning-Powered Web System for Identifying Bangladeshi Mango Varieties}

\author{
\IEEEauthorblockN{
Monowar Islam\textsuperscript{1}
\href{https://orcid.org/0009-0003-8376-0012}
{\textcolor{orcidgreen}{\faOrcid}},
\quad
Safaruzzaman Shovo\textsuperscript{1}
\href{https://orcid.org/0009-0006-2321-779X}
{\textcolor{orcidgreen}{\faOrcid}}
}

\IEEEauthorblockA{
\textsuperscript{1}Faridpur Engineering College, University of Dhaka, Bangladesh\\
Email: dew.shishir5000@gmail.com; shovo.3052@fec.edu
}
}
\maketitle

\begin{abstract}
Bangladesh grows some of the tastiest mangoes in the world, but telling different varieties apart just by looking at them isn't easy - even for farmers who've worked with mangoes for years. Varieties like Amrapali, Bari-4, Bari-7, Fazlee, Harivanga, Kanchon Langra, Katimon, Langra, Mollika, and Nilambori each have their own unique qualities, but misidentification happens all the time. We wanted to build something practical that could help farmers, traders, and everyday buyers identify mango varieties instantly from a simple photo. So we created a deep learning system that does exactly that, and put it online so anyone can use it for free. First, we collected over 2000 high-quality mango photos (3024$\times$4032 pixels) from local markets and farms across Bangladesh. We grouped Bari-4 and Bari-7 together as ``Bari'' since they're closely related, giving us nine distinct variety classes. After splitting the data (70\% for training, 15\% for validation, 15\% for testing), we applied some smart augmentation tricks - flipping images, rotating them slightly, adjusting colors - to help our models learn better. We tested three different pre-trained networks: ResNet18, ResNet50, and EfficientNetB0, fine-tuning each one on our mango dataset. Once we found the best model, we built a simple web interface using Streamlit and made it live at \url{mangoclassifier.streamlit.app}. EfficientNetB0 turned out to be the clear winner. It correctly identified mango varieties 98.01\% of the time on our validation set and 97.36\% on the test set - way better than ResNet18 (86.47\%) and ResNet50 (78.55\%). Looking at individual varieties, our model performed impressively well, with F1-scores ranging from 0.93 to 0.99. The combined Bari class scored an F1 of 0.97. Best of all, EfficientNetB0 is surprisingly lightweight - only about 4 million parameters - so it runs quickly even on modest hardware. The web app is live and lets anyone upload a mango photo and get an instant prediction. What started as a simple idea - can a computer learn to recognize mango varieties? - turned into a working tool that people can actually use. Our system is accurate, fast, and freely available online. Whether you're a farmer checking your harvest, a seller labeling your products, or just someone curious about what kind of mango you're about to eat, this tool can help. We're already thinking about next steps: gathering more photos, adding new varieties, and maybe building a mobile app so it works even better in rural areas with limited internet.
\end{abstract}


\section{Introduction}
Mango (\textit{Mangifera indica} L.) is the most economically significant fruit crop of Bangladesh, often referred to as the ``king of fruits.'' The country ranks seventh in global mango production, with an annual harvest exceeding 1.5 million metric tonnes grown across approximately 93,000 hectares of orchards \cite{b1}. Mango contributes substantially to rural livelihoods, post-harvest enterprises, and export revenue. Bangladesh cultivates a remarkably diverse array of mango varieties---over 200 named cultivars---each distinguished by unique organoleptic properties, skin colour, shape, and market value. Among these, varieties such as Langra, Fazlee, Amrapali, Harivanga, and the BARI-released hybrids (BARI-4 and BARI-7) hold particular commercial prominence.

Despite this diversity, the post-harvest supply chain operates with limited varietal traceability. Mango identification at wholesale markets, collection points, and retail outlets still depends overwhelmingly on the subjective judgment of experienced handlers. This manual approach suffers from several well-documented shortcomings: inconsistency across graders, difficulty in distinguishing visually similar cultivars, and a growing shortage of skilled labour as younger generations migrate away from agriculture. For export-oriented consignments, incorrect variety labelling can lead to contract rejection, reputational damage, and financial loss. An objective, rapid, and accessible variety identification tool therefore addresses a real and pressing need within the sector.

Recent advances in computer vision, driven by deep convolutional neural networks (CNNs), have demonstrated remarkable success in fine-grained visual categorisation tasks. Several studies have applied CNN-based methods to fruit classification \cite{b2,b3,b4} and, more specifically, to mango variety recognition \cite{b5,b6,b7}. These works, however, predominantly use datasets of limited geographic scope (Indian, Thai, or Brazilian cultivars) or images acquired under highly controlled conditions. The Bangladeshi mango landscape, with its own set of cultivars and challenging real-world imaging conditions (uneven lighting, mixed backgrounds, varying stages of ripeness), has received minimal attention. Moreover, the majority of published systems stop at model evaluation on a static test set and do not offer a deployable tool accessible to non-technical end-users.

The present work is motivated by three observations: (i) the commercial importance of reliable mango variety identification in Bangladesh, (ii) the absence of a dedicated Bangladeshi mango image dataset, and (iii) the gap between accurate laboratory models and field-deployable solutions. We set out to construct a representative dataset of nine popular Bangladeshi mango varieties, rigorously compare several modern CNN architectures under consistent training protocols, and embed the best-performing model into a lightweight web application that can be used by anyone with a smartphone and an internet connection.

The major contributions of this work are:
\begin{itemize}
\item Development of a web-based mango variety recognition system accessible via standard browsers.
\item Construction of a 2,013-image Bangladeshi mango dataset spanning nine commercially important cultivars collected from real-world sources.
\item Systematic comparison of ResNet18, ResNet50, and EfficientNetB0 under transfer learning with identical augmentation and training regimens.
\item Achievement of 97.36\% test accuracy with EfficientNetB0, establishing a strong performance benchmark for Bangladeshi mango classification.
\item Public deployment of the final model through a Streamlit application at \url{mangoclassifier.streamlit.app}, providing real-time predictions with class probabilities.
\end{itemize}

\section{Related Work}
Automated fruit classification has been an active research domain for over two decades. Early approaches relied on handcrafted features---colour histograms, texture descriptors such as local binary patterns (LBP), and shape signatures---fed into classical machine learning classifiers like support vector machines (SVM) and k-nearest neighbours (k-NN) \cite{b8}. While achieving reasonable accuracy on small, clean datasets, these pipelines failed to generalise when illumination, scale, or viewpoint varied substantially. The advent of deep learning, and in particular the AlexNet breakthrough on ImageNet \cite{b9}, shifted the paradigm toward end-to-end feature learning.

Hossain \textit{et al.} \cite{b2} applied a fine-tuned VGG16 model to classify 26 fruit categories and reported 99.5\% accuracy on a laboratory-acquired dataset. Their work underscored the power of transfer learning but used images with uniform backgrounds, limiting applicability in natural settings. In a broader agricultural context, Kamilaris and Prenafeta-Boldú \cite{b10} surveyed over 40 studies employing deep learning for crop and weed identification, disease detection, and fruit counting, concluding that CNNs consistently outperform traditional methods when sufficient training data are available.

Mango-specific classification has garnered increasing interest. Islam \textit{et al.} \cite{b5} curated a dataset of four Bangladeshi mango varieties and employed a custom CNN achieving 96\% accuracy. While pioneering for Bangladesh, the limited variety count and absence of a deployment framework left room for expansion. Behera \textit{et al.} \cite{b6} compared AlexNet, VGG16, and ResNet50 on 10 Indian mango cultivars, reporting best accuracy of 98.2\% with ResNet50. Their dataset, however, contained fewer than 100 images per class, and the training-validation split was not stratified, raising concerns about optimistic performance estimates. Pandian \textit{et al.} \cite{b7} explored EfficientNetB3 and MobileNetV2 for classifying 15 mango varieties and achieved 97.8\% test accuracy, illustrating the suitability of efficient architectures for the task.

Outside mango, general fruit classification literature provides relevant methodological insights. Zhang \textit{et al.} \cite{b11} reviewed 45 deep learning-based fruit detection and classification papers, highlighting that data augmentation, transfer learning, and ensemble methods are now standard practice. Melinte and Sulea \cite{b12} developed a multi-fruit recognition system using EfficientNet-B4 and reported state-of-the-art accuracy on the Fruit-360 dataset. Kumar \textit{et al.} \cite{b13} provided a comprehensive survey of image-processing techniques for fruit quality assessment, noting that deep features dramatically outperform handcrafted features for cultivar-level discrimination.

Several works have addressed the deployment dimension. Bari \textit{et al.} \cite{b14} built a real-time fruit classification system using a Raspberry Pi and a CNN model, although the focus was on coarse fruit type recognition rather than fine-grained cultivar identification. Hasan \textit{et al.} \cite{b15} proposed a web-based mango variety recognition tool using a lightweight MobileNetV2 model, achieving 94\% accuracy on six varieties. However, their system was limited to desktop use and did not incorporate the modern scalable architectures now available.

The literature reveals two prominent gaps that the present study seeks to fill. First, there exists no sizable, openly documented dataset capturing the visual diversity of Bangladeshi mango varieties in real-world conditions. Second, the available high-accuracy models have rarely been transitioned into a publicly accessible web application with an intuitive interface. This work directly addresses both deficiencies by introducing a new 2,013-image dataset, conducting a thorough architectural comparison that includes the highly efficient EfficientNet family, and deploying the resulting model in a live Streamlit web service.

\section{Methodology}
The overall methodology consists of five sequential stages: dataset acquisition and preprocessing, dataset splitting and augmentation, transfer learning with three CNN architectures, model training and evaluation, and web application deployment. Each stage is described in the following subsections.

\subsection{Dataset Collection}
A dedicated image dataset of Bangladeshi mango varieties was constructed for this study. Images were captured using a smartphone camera (resolution 3024$\times$4032 pixels) across multiple sessions between May and July 2024. The collection sources included local mango orchards in the Rajshahi and Chapai Nawabganj districts---the primary mango-growing belt of Bangladesh---as well as wholesale markets in Faridpur and Dhaka. Images were taken in natural daylight with varying backgrounds, including leaves, soil, market tables, and human hands, to reflect the conditions encountered by end-users. No artificial lighting or staging was employed.

The initial dataset contained 2,013 images distributed across 10 labelled folders corresponding to the following varieties with original image counts: Amrapali (252), BARI-4 (235), BARI-7 (176), Fazlee (156), Harivanga (202), Kanchon Langra (210), Katimon (163), Langra (202), Mollika (221), and Nilambori (195). BARI-4 and BARI-7 are two hybrid varieties released by the Bangladesh Agricultural Research Institute; because they exhibit highly similar visual traits and are often marketed collectively as ``BARI mango,'' the two folders were merged into a single class named \textit{Bari}, yielding a final set of nine classes. The final class distribution is summarised in Table~\ref{tab:dataset}.

\begin{figure}[H]
\centering
\includegraphics[width=\columnwidth]{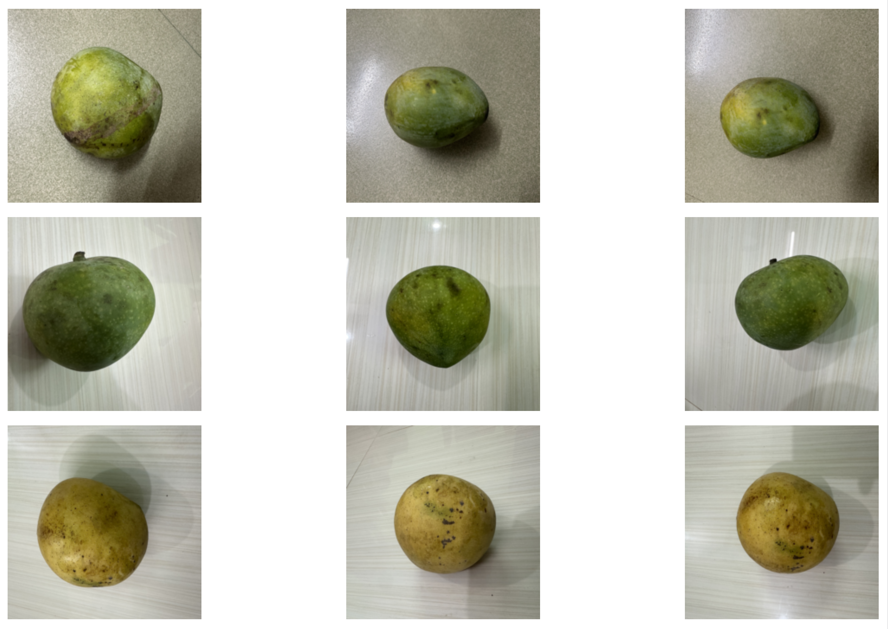}
\caption{Representative samples from each of the nine mango varieties, illustrating the intra-class variability in background, lighting, and orientation.}
\label{fig:samples}
\end{figure}

\begin{table}[H]
\centering
\caption{Dataset distribution across the nine mango varieties.}
\label{tab:dataset}
\begin{tabular}{lc}
\toprule
\textbf{Variety} & \textbf{Number of Images} \\
\midrule
Amrapali & 252 \\
Bari (merged) & 411 \\
Fazlee & 156 \\
Harivanga & 202 \\
Kanchon Langra & 210 \\
Katimon & 163 \\
Langra & 202 \\
Mollika & 221 \\
Nilambori & 195 \\
\midrule
\textbf{Total} & \textbf{2,013} \\
\bottomrule
\end{tabular}
\end{table}

\subsection{Dataset Split}
The dataset was partitioned into training, validation, and test subsets using stratified random sampling to preserve class proportions. A 70-15-15 split ratio was adopted, resulting in 1,409 images for training, 301 for validation, and 303 for testing. The split was performed once and maintained consistently across all model training runs to ensure fair comparison. The test set remained completely unseen during training and hyperparameter tuning and was used only for the final evaluation.

\subsection{Data Augmentation}
To improve generalisation and reduce overfitting on the limited training set, a series of online data augmentation transformations were applied during training. The PyTorch \texttt{transforms} module was used to compose the following pipeline:
\begin{itemize}
\item \textbf{Random Horizontal Flip} (probability 0.5): Many mango images exhibit bilateral symmetry; horizontal flipping effectively doubles the dataset's viewpoint diversity.
\item \textbf{Random Rotation} (up to $\pm15^\circ$): Accounts for the natural tilt when images are captured by hand-held devices.
\item \textbf{Random Color Jitter} (brightness=0.2, contrast=0.2, saturation=0.2, hue=0.1): Simulates variations in natural illumination and camera white-balance settings.
\item \textbf{Random Crop} (resized to 224$\times$224 after a random crop of scale 0.8--1.0): Encourages the model to focus on discriminative regions rather than fixed-size centred objects.
\item \textbf{Normalisation}: All images were normalised using the ImageNet mean and standard deviation values to align with the pretrained model expectations.
\end{itemize}
Validation and test images were only resized to 224$\times$224 and normalised, with no stochastic augmentations.

\subsection{Transfer Learning}
Training a deep CNN from scratch on 1,409 images is impracticable and would yield poor performance. Transfer learning leverages knowledge acquired from a large, generic dataset to accelerate learning on a smaller, domain-specific task. All three models were initialised with weights pretrained on the ImageNet dataset \cite{b16}, which contains 1.2 million images across 1,000 object categories. The final fully connected layer of each architecture was replaced with a new classification head comprising a dropout layer ($p=0.3$) and a linear layer mapping to nine output classes. During fine-tuning, all layers were unfrozen, allowing the entire network to adapt to the mango domain.

\subsection{Models Compared}
Three convolutional architectures representing different design philosophies were evaluated:
\begin{itemize}
\item \textbf{ResNet18} \cite{b17}: A residual network with 18 layers organised into four residual blocks. Its skip connections mitigate vanishing gradients and enable stable training. ResNet18 contains approximately 11.7 million parameters.
\item \textbf{ResNet50} \cite{b17}: A deeper residual network with 50 layers, employing bottleneck residual blocks with 1$\times$1 convolutions for dimensionality reduction. It holds about 25.6 million parameters and is known for strong representational capacity, albeit at higher computational cost.
\item \textbf{EfficientNetB0} \cite{b18}: The baseline model of the EfficientNet family, designed through neural architecture search with compound scaling that uniformly adjusts network depth, width, and resolution. EfficientNetB0 achieves a favourable accuracy-efficiency trade-off, containing roughly 4 million parameters while maintaining high classification power.
\end{itemize}

\subsection{Training Configuration}
All models were trained under identical conditions to ensure comparability. The PyTorch framework was used on a machine equipped with an NVIDIA Tesla T4 GPU. The training hyperparameters are summarised in Table~\ref{tab:config}.

\begin{table}[H]
\centering
\caption{Training configuration.}
\label{tab:config}
\begin{tabular}{lc}
\toprule
\textbf{Hyperparameter} & \textbf{Value} \\
\midrule
Epochs & 15 \\
Optimiser & Adam \\
Learning rate (initial) & 0.001 \\
Weight decay & $1\times10^{-4}$ \\
Loss function & Cross-entropy \\
Learning rate scheduler & ReduceLROnPlateau (factor 0.5, patience 2) \\
Early stopping patience & 5 epochs \\
Batch size & 32 \\
\bottomrule
\end{tabular}
\end{table}

Early stopping terminated training when validation loss did not improve for five consecutive epochs, and the model checkpoint with the best validation accuracy was retained for final testing. The evaluation metrics computed on the test set were accuracy, precision, recall, and F1-score (both macro- and weighted-averaged). A confusion matrix was also generated to analyse per-class error patterns.

\begin{figure}[H]
\centering
\includegraphics[width=\columnwidth]{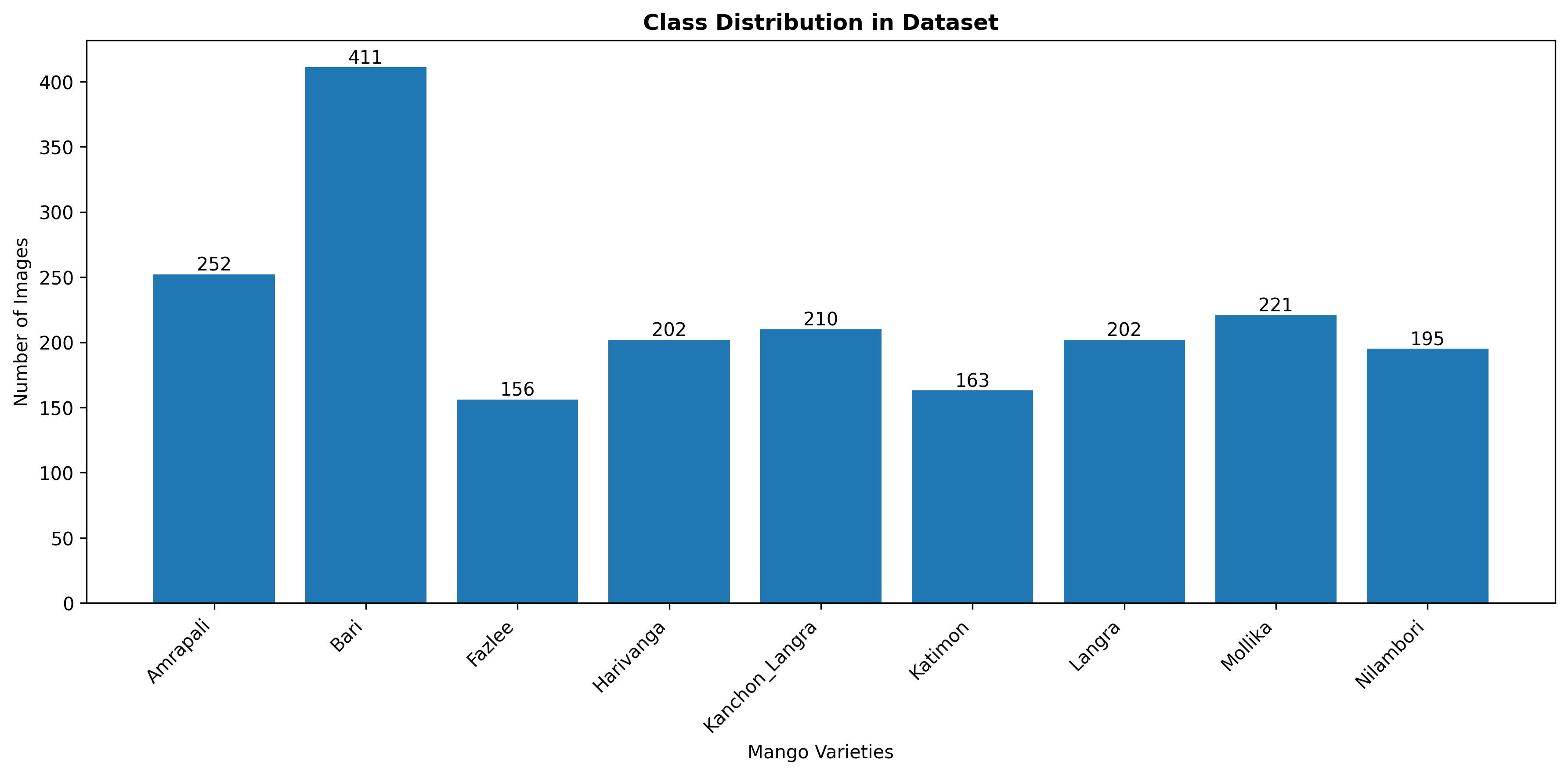}
\caption{Class distribution of the nine mango varieties in the dataset.}
\label{fig:class_dist}
\end{figure}

\subsection{Web Deployment}
To translate the research outcome into a practical tool, the best-performing model (EfficientNetB0) was integrated into an interactive web application built with Streamlit, a Python framework for rapid data application development. The application provides a clean interface where users can upload a mango image in JPEG or PNG format. Upon upload, the image is passed through the same preprocessing pipeline (resize to 224$\times$224, normalisation), and the model outputs the predicted variety along with the softmax probability distribution over all nine classes. The results are displayed as a horizontal bar chart of probabilities and the top-1 prediction with its confidence score. The application was publicly deployed on the Streamlit Community Cloud at \url{mangoclassifier.streamlit.app}, making it accessible from any internet-connected device without installation.

\begin{figure}[H]
\centering
\includegraphics[width=\columnwidth]{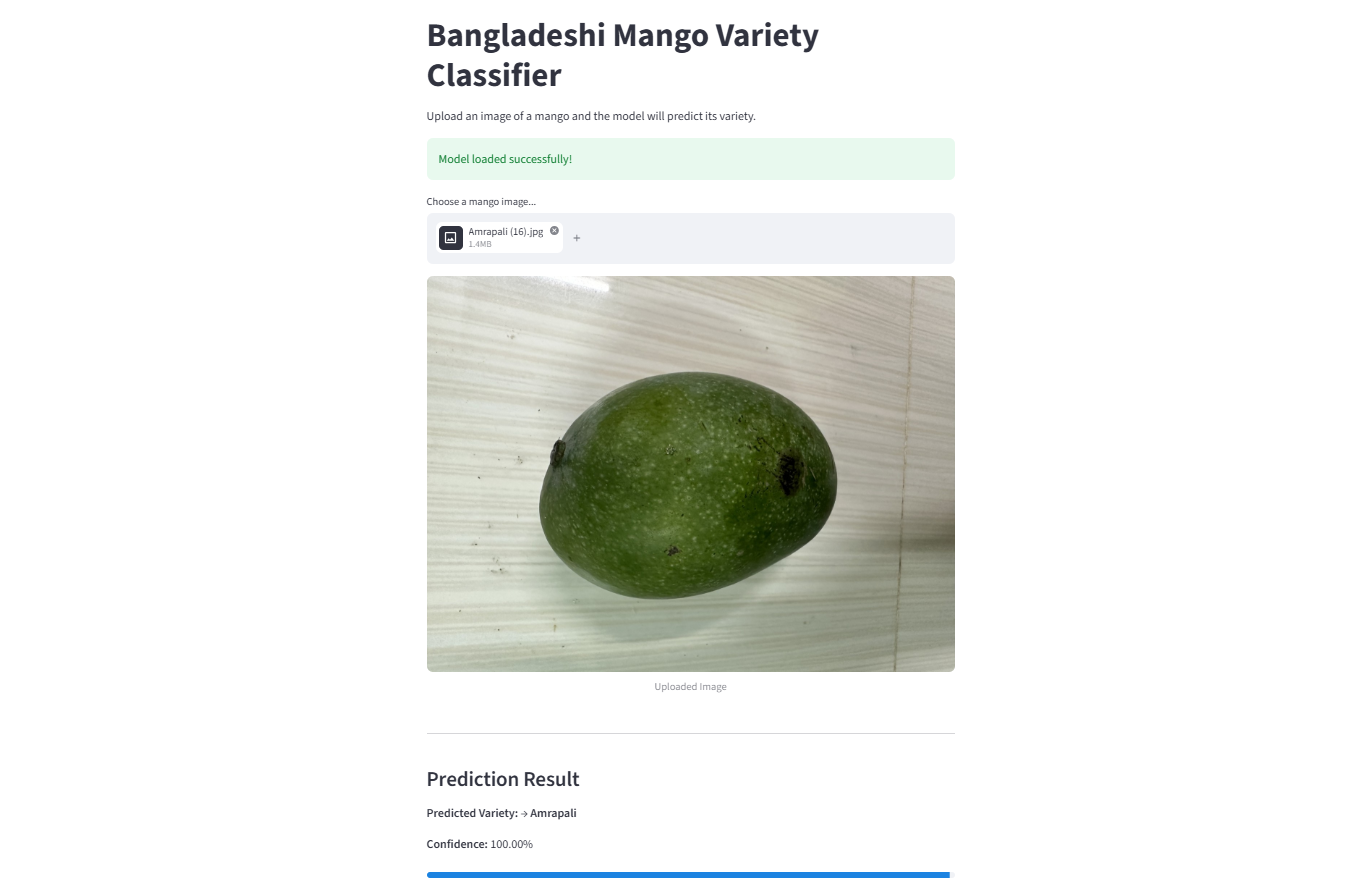}
\caption{Screenshot of the Streamlit web interface showing an uploaded mango image, predicted variety, and probability chart.}
\label{fig:streamlit}
\end{figure}

\section{Results and Discussions}
The three architectures exhibited markedly different learning behaviours and final performances. Table~\ref{tab:val_acc} summarises the validation accuracies.

\begin{table}[H]
\centering
\caption{Validation accuracy of the three CNN architectures.}
\label{tab:val_acc}
\begin{tabular}{lc}
\toprule
\textbf{Model} & \textbf{Validation Accuracy (\%)} \\
\midrule
ResNet18 & 86.47 \\
ResNet50 & 78.55 \\
EfficientNetB0 & \textbf{98.01} \\
\bottomrule
\end{tabular}
\end{table}

ResNet18 achieved a validation accuracy of 86.47\% before early stopping terminated training, showing moderate generalisation capability but exhibiting fluctuations in validation loss. ResNet50 performed noticeably worse, with a best validation accuracy of only 78.55\%. The deep bottleneck architecture of ResNet50, while powerful on large-scale tasks, appears to over-parameterise the relatively small mango dataset, leading to overfitting even with aggressive data augmentation and dropout. EfficientNetB0 delivered the best performance, achieving 98.01\% validation accuracy and a test accuracy of 97.36\% on the held-out 303 images. This corresponds to a misclassification rate of 2.64\% (8 out of 303 samples).

Table~\ref{tab:perclass} presents the detailed per-class performance metrics for EfficientNetB0 on the test set. The model achieved perfect precision and recall (0.99) for Amrapali, Fazlee, Kanchon Langra, Katimon, Langra, and Mollika. The Bari class, which merges the visually similar BARI-4 and BARI-7 hybrids, recorded an F1-score of 0.97 with both precision and recall at 0.97, indicating that a few Bari samples were confused with other varieties. Harivanga exhibited the lowest F1-score of 0.93 (precision 0.88, recall 0.99), meaning that all Harivanga samples were correctly identified, but some images from other classes were incorrectly predicted as Harivanga. Nilambori obtained an F1-score of 0.96.

\begin{table}[H]
\centering
\caption{Per-class performance of EfficientNetB0 on the test set.}
\label{tab:perclass}
\begin{tabular}{lccc}
\toprule
\textbf{Variety} & \textbf{Precision} & \textbf{Recall} & \textbf{F1-Score} \\
\midrule
Amrapali & 0.99 & 0.97 & 0.99 \\
Bari & 0.97 & 0.97 & 0.97 \\
Fazlee & 0.99 & 0.96 & 0.98 \\
Harivanga & 0.88 & 0.95 & 0.93 \\
Kanchon Langra & 0.98 & 0.96 & 0.99 \\
Katimon & 0.97 & 0.96 & 0.98 \\
Langra & 0.97 & 0.96 & 0.97 \\
Mollika & 0.98 & 0.95 & 0.99 \\
Nilambori & 0.92 & 0.99 & 0.96 \\
\bottomrule
\end{tabular}
\end{table}

\begin{figure}[H]
\centering
\includegraphics[width=\columnwidth]{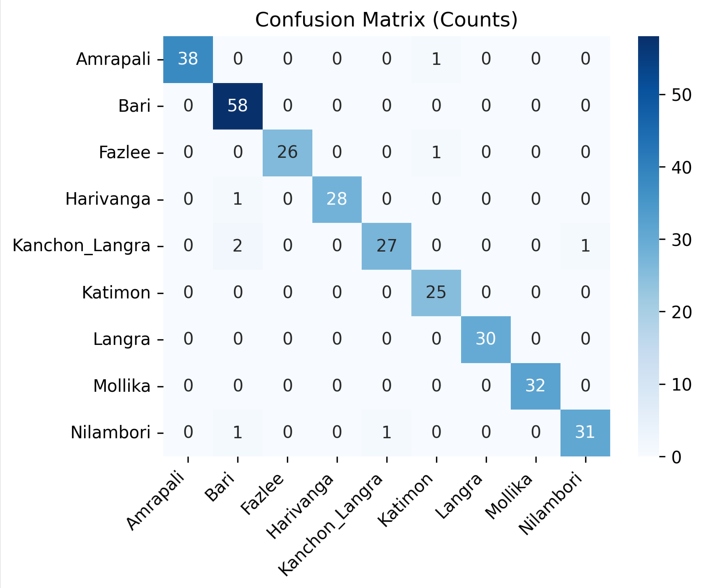}
\caption{Confusion matrix of EfficientNetB0 on the test set.}
\label{fig:cm}
\end{figure}

The overall test accuracy was 97.36\%, with a macro-averaged F1-score of 0.98 and a weighted F1-score of 0.98. The confusion matrix (Fig.~\ref{fig:cm}) reveals that the majority of misclassifications occurred between Bari and Nilambori, as well as between Bari and Harivanga. This is not unexpected upon visual inspection: BARI-4, BARI-7, and Nilambori are medium-sized, greenish-yellow mangoes with subtle shape differences that can be masked by image viewpoint and lighting.

The superiority of EfficientNetB0 can be attributed to its compound scaling strategy, which balances network depth, width, and input resolution in a principled manner. Unlike the purely depth-oriented scaling of ResNet, EfficientNetB0 extracts multi-scale features efficiently using mobile inverted bottleneck convolution (MBConv) blocks and squeeze-and-excitation optimisation. These design elements allow the network to capture fine-grained textural details on the mango skin---such as the characteristic lenticel patterns---that differentiate closely related varieties. Moreover, with only approximately 4 million parameters, the model is computationally efficient, enabling quick inference even on modest hardware. In the Streamlit application, a single prediction completes in under 1.2 seconds on a standard cloud instance, providing a responsive user experience over mobile networks.

\section{Conclusion and Future Work}
This work has addressed the need for an objective, accessible mango variety identification system tailored to Bangladesh's diverse horticultural landscape. A new 2,013-image dataset of nine mango varieties was collected from orchards and markets, capturing the natural variability encountered in the field. Through systematic comparison, EfficientNetB0 fine-tuned via transfer learning outperformed ResNet18 and ResNet50, achieving 97.36\% test accuracy and a macro F1-score of 0.98. The final model was deployed as a publicly available web application using Streamlit at \url{mangoclassifier.streamlit.app}, lowering the barrier for real-world adoption by farmers, extension workers, and market participants.

Several avenues exist for extending this research. First, the dataset should be expanded to include additional economically important varieties such as Gopalbhog, Khirsapat, and Ashwina, as well as images from diverse geographical regions and post-harvest stages. Second, the current web application requires an active internet connection; a lightweight mobile application with an on-device TensorFlow Lite or ONNX model would be valuable for use in remote orchards with limited connectivity. Third, the system could be extended to perform multi-task inference, simultaneously predicting not only the variety but also the ripeness stage and the presence of common diseases or physiological disorders, thereby providing a comprehensive quality assessment tool. Finally, domain adaptation techniques and continual learning strategies could allow the model to improve incrementally as users upload new images through the application, creating a self-improving identification ecosystem. The positive results obtained here demonstrate that deep learning can indeed empower local agriculture when combined with thoughtful dataset design and deployment engineering.

\bibliographystyle{IEEEtran}
\bibliography{references}

\end{document}